\documentclass[letterpaper, 10 pt, conference]{ieeeconf}  

\IEEEoverridecommandlockouts                              

\title{\LARGE \bf
Semi-Automated Nasal PAP Mask Sizing using Facial Photographs}

\author{Benjamin Johnston, Alistair McEwan and Philip de Chazal
\thanks{Research supported by ARC grant FT110101098}
\thanks{Ben Johnston and Philip de Chazal are with the Sleep Research Group, Charles Perkins Centre, School of Electrical and Information Engineering, University of Sydney, Sydney, NSW, 2006, Australia, (phone: +612 911 41528; e-mails: \tt\small{\{ben.johnston, philip.dechazal\}@sydney.edu.au)}}%
\thanks{Alistair McEwan is with the School of Electrical and Information Engineering, University of Sydney, (e-mail: \tt\small{alistair.mcewan@sydney.edu.au)}}%
}
\usepackage{amsmath, graphicx, wrapfig, verbatim, amssymb, wasysym, multirow}

\begin{document}
%
\maketitle
\begin{abstract}
We present a semi-automated system for sizing nasal Positive Airway Pressure (PAP) masks based upon a neural network model that was trained with facial photographs of both PAP mask users and non-users. It demonstrated an accuracy of 72\% in correctly sizing a mask and 96\% accuracy sizing to within 1 mask size group. The semi-automated system performed comparably to sizing from manual measurements taken from the same images which produced 89\% and 100\% accuracy respectively.

\end{abstract}
\begin{keywords}
OSA, PAP, CPAP, neural networks, machine learning, facial landmarking, telemedicine, sizing
\end{keywords}
\section{Introduction}
\label{sec:intro}

Positive airway pressure or continuous positive airway pressure (CPAP) therapy, seminally described by Sullivan et al. in 1981 \cite{Sullivan1981} is generally accepted as the gold standard treatment for obstructive sleep apnoea (OSA) \cite{AmThor1994}.  Positive airway pressure effectively treats OSA by providing a pneumatic splint at the location of soft palate, preventing collapse onto the pharynx and thus allowing the passage of air to the lungs.  PAP therapy is applied using a specially manufactured device that consists 3 of main components: a pressure generator, which is a highly sophisticated air pump used to compress atmospheric air for delivery into the pharynx; a mask which the patient wears over the nose and / or mouth to ensure the pressurised air is delivered to the correct location and finally a humidifier unit. 

The mask is a critical component of the PAP system as it is the device that is in intimate contact with the patient and must be capable of maintaining treatment pressure for the duration of a night's sleep.  Other than potentially reducing treatment efficacy, issues with mask performance are often immediately noticeable for the patient such as uncomfortable air leaks into the eye or visible red marks on the face caused by mask rubbing during use.  While incidents of side effects due to mask usage vary, a number of studies have observed that many patients experience difficulties using their mask \cite{Berry2000}. Amfilochiou et al. reported incidences of mask leak and red marks in 48\% and 40\% of participants \cite{Amfilochiou2009}; Gay et al. noted that mask interface issues are the most commonly reported PAP related side effects \cite{Gay2006}.  

In the context of PAP therapy compliance, incidents of device related side-effects such as mask leakage are not inconsequential.  Engleman in 1994 observed that patients who experienced side effects used their PAP devices less that those who did not \cite{Engleman1994}.  More recently in 2003 Massie and Hart, while studying differences between nasal and nasal pillows mask types noted a negative correlation with the occurrence of air leaks and sore eyes with CPAP usage \cite{Massie2003}.  These side effects often form a barrier to effective therapy usage, particularly with new patients \cite{Dickerson2006} being introduced to PAP for the first time.  It has been found that the first week of therapy is critical in determining long term therapy use patterns, as it is during this time a patient decides if they are going to continue or abandon treatment \cite{Weaver1997, Stepnowsky2002}.  Given these characteristics and that patients who swap their mask due to poor fit or comfort problems are 7 times more likely to discontinue therapy \cite{Bachour2016} it is critical that patients are issued with the correct mask size and type on the first attempt.

\subsection{Current Mask Sizing}
\label{sec:current_sizing}
While the exact details of sizing PAP masks varies between each of the manufacturers, there are two common methods which are generally used amongst clinicians.  One process uses a fitting template provided by the manufacturer such as that illustrated in Figure \ref{fig:datasample}a.  When using these templates the patient places their nose within specific features of the template and the clinician matches the appropriate size with indications on the guide.  The commonly used alternative sizing methodology uses the experience and expertise of the clinician to make the mask choice without the use of a guide.   This paper describes the potential use of a third, automated method of PAP mask sizing with the aim of improving sizing accuracy and reducing mask related side effects.

\subsection{Potential Applications of Automated Sizing}
\label{sec:applications}
Automated PAP mask sizing could provide significant benefit to new patients requiring PAP therapy who live in remote or rural communities and have limited access to medical services.  Through the use of Home Sleep Tests (HST) an OSA diagnosis can be issued without the patient leaving the home.  However in order to issue a PAP system with appropriate mask an in-person consult (possibly requiring hours of travel) is currently required.  An automated sizing system could enable the use of a \textit{telemedicine} approach to PAP consults.  A patient would send a facial photograph to a clinician who would use software to analyse the photograph and automatically determine the correct mask size. The mask would then be sent to the patient, thus avoiding an in-person visit.  Such a system could also provide additional benefit by reducing consult time and increasing the capacity of a clinician to see more patients.  In this study we have developed a semi-automated PAP mask sizing system by processing facial images.

\section{Image Dataset}
\label{sec:image_dataset}
\begin{figure}[h]
\begin{minipage}[t]{.48\linewidth}
  \centering
  \centerline{\includegraphics[height=3.5cm]{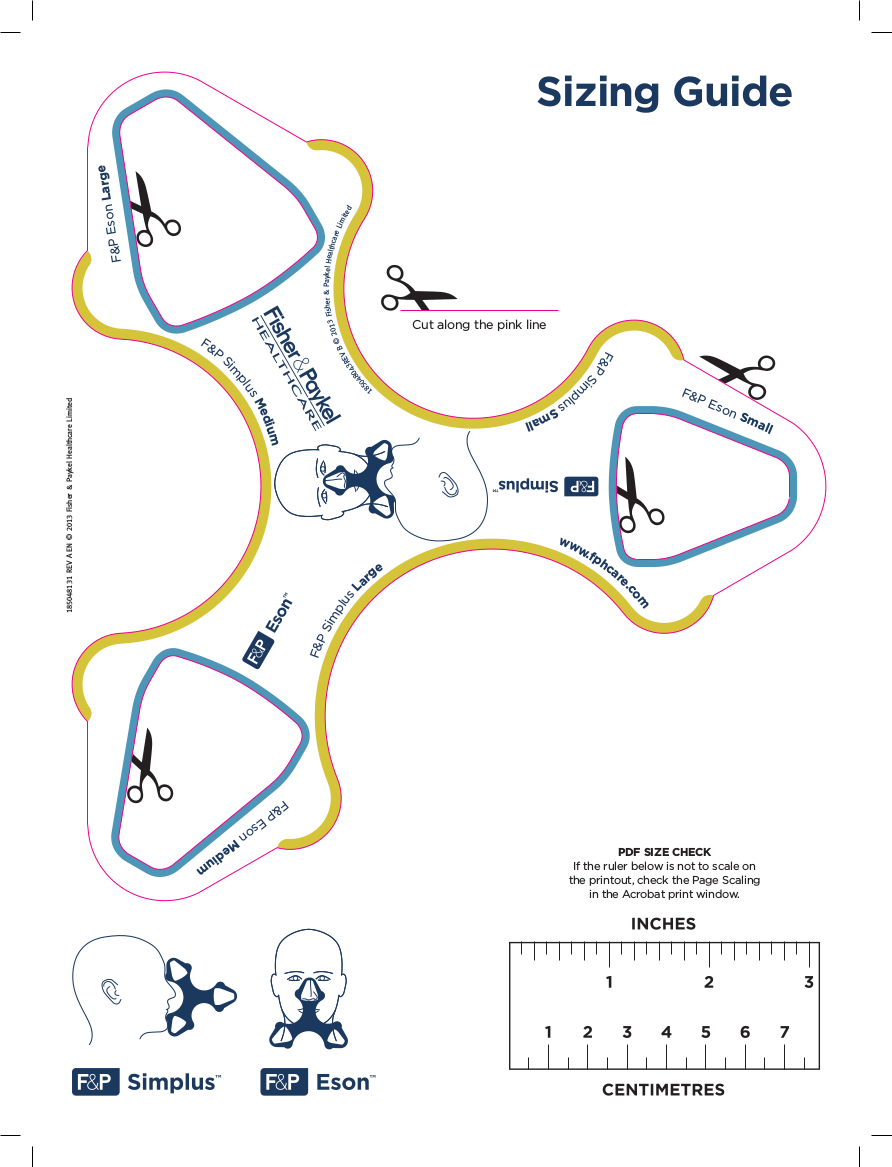}}
  
  \centerline{(a) Eson sizing template}\medskip
  
\end{minipage}
\hfill
\begin{minipage}[t]{0.48\linewidth}
  \centering
  \centerline{\includegraphics[height=3.5cm]{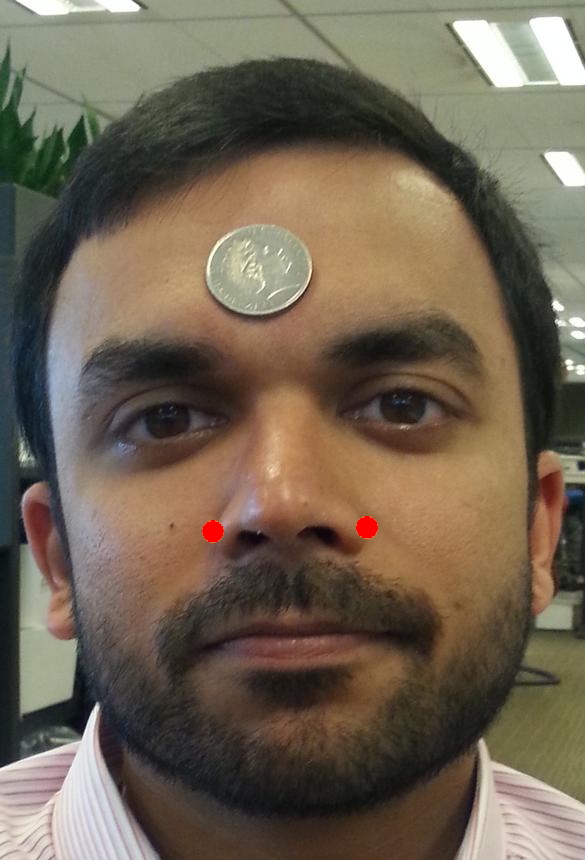}}
  \centerline{(b) Labelled image}\medskip
\end{minipage}

\begin{minipage}[b]{1\linewidth}
  \centering
  \centerline{\includegraphics[height=2.8cm]{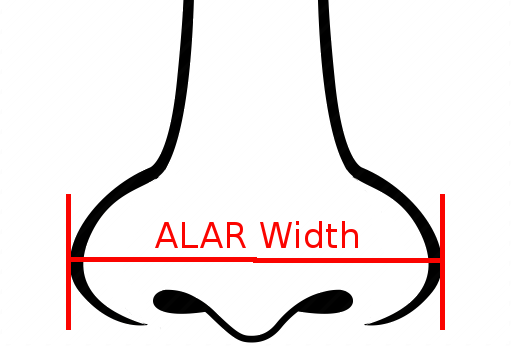}}
  
  \centerline{(c) Illustration indicating ALAR Width}\medskip
  
\end{minipage}

\caption{Dataset (sample image used with permission from Priyanshu Gupta)}
\label{fig:datasample}
\end{figure}

The dataset used throughout this study comprised 251 facial images of male and female, PAP (49 samples) and non-PAP users that were collected in a variety of locations using smartphone cameras.  Care was taken during photography to ensure the face was not obscured by glasses, hair or hats and that the pose of the face was approximately parallel with the smartphone.  Participants had their the ALAR or nose width (see Figure \ref{fig:datasample}c) measured at the time of image capture using vernier callipers. These nose width measurements were applied against a sizing template for an Eson nasal PAP mask manufactured by Fisher \& Paykel (F\&P) Healthcare (Figure \ref{fig:datasample}a \& Table \ref{tab:eson_sizes}) to determine the `ground truth' mask size for each of the selected participants.

Each image in the set (e.g. Figure \ref{fig:datasample}b) is a frontal photographic shot of the participant with an Australian 20 cent piece placed on the forehead to provide an indication of scale.

\section{Methods}
\label{sec:methods}

Following data collection, the face from each image was extracted using the opencv implementation of the Viola-Jones algorithm\cite{Viola2001} into a sub-image and resized to 512 x 512 pixels.  The pixel locations of the left and right lateral nasal walls were then manually identified and recorded for each image. The scale of each image was recorded in pixels / mm using ImageJ (produced by the US National Institute of Health)  by measuring the number pixels that span the diameter and dividing by the known diameter of the coin (28.65mm).

\begin{table}[b]
\renewcommand{\arraystretch}{1.2}
\begin{tabular}{| c | c | c |}
	\hline
	\textbf{Size} & \textbf{Lower Limit (mm)} & \textbf{Upper Limit (mm)} \\ \hline
	Small & 0 & 37 \\ \hline
	Medium & 37 & 41 \\ \hline
	Large & 41 & 45 \\ \hline
	Too Large* & 45 & $\infty$ \\ \hline
\end{tabular}

\caption{Fisher \& Paykel Eson mask sizes. *Not actually a size but is used to represent individuals who do not fit an Eson mask}
\label{tab:eson_sizes}
\end{table}

\subsection{Size Overlap}
\label{sec:size_overlap}

The PAP mask industry provides a margin of overlap for mask sizes as the means of determining the correct mask size is not an exact science (see Section \ref{sec:current_sizing}). Those mask users who fall within the boundary between two sizes will achieve acceptable mask performance with either of the adjacent mask sizes. This margin of acceptable error provides some tolerance to mis-sizing. 
To reflect this situation in the field, we applied a 2\% tolerance to sizing decisions within this study. For individuals whose `ground truth' nose width fell within 2\% of a size boundary e.g 36.26 - 37.74mm for a small / medium mask, the predicted mask size was considered correct if either a small or medium mask is predicted.

\subsection{Data Pre-Processing}
\label{sec:pre-processing}

\begin{figure*}[!t]

\begin{minipage}[b]{1\linewidth}
  \centering
  \centerline{\includegraphics[height=4.1cm]{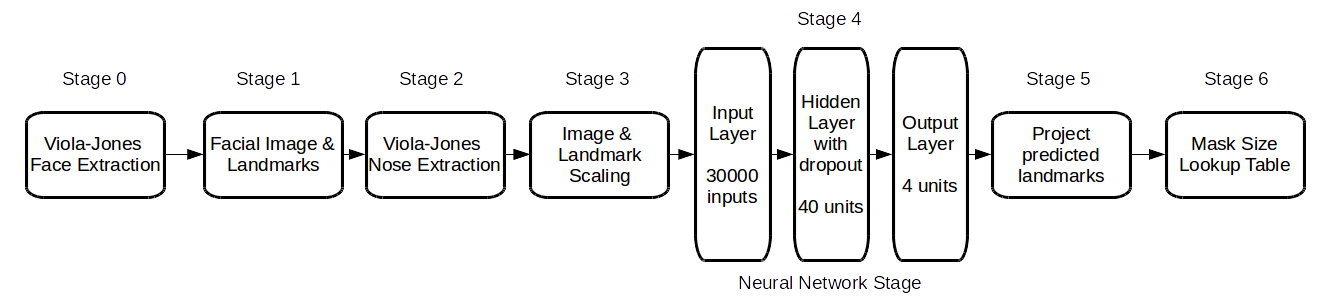}}
\end{minipage}
\caption{Processing stages of the model}
\label{fig:model_stages}
\end{figure*}

As nasal mask sizes are solely determined by the width of a user's nose, the nasal region of each image was extracted in grayscale using the opencv implementation of the Viola-Jones algorithm\cite{ Viola2001} and the pre-trained nose haarcascade classifier provided with opencv (Stages 1 \& 2 of Figure \ref{fig:model_stages}).  Each $200\times150$ image was unwrapped to form a data vector $(\mathbf{x}\in\mathbb{R}^{251 \times 30000})$ and the two nasal landmark coordinates were unwrapped to give a 4 element target vector $(\mathbf{x}\in\mathbb{R}^{251 \times 4})$.  The $x$ and $y$ datasets were scaled and centered between -1 and 1 by dividing by their respective maxima and subtracting the resulting mean (Stage 3 Figure \ref{fig:model_stages}).

\subsection{Manual Mask Sizing from the Images}
\label{sec:man_mask_size_images}
In an attempt to determine the optimum performance that could be achieved by the system, mask sizes were also calculated manually using the labelled landmarks of the test set facial photographs.  The physical nose width was calculated by dividing by the distance between the two labelled landmarks (pixels) and the scale (pixels / mm) as provided by the 20 cent piece.  This physical measurement was compared with Table \ref{tab:eson_sizes} to determine the sizes referred to as \textit{Manually Measured} throughout this study.

\subsection{Neural Network Architecture}
\label{sec:neural_net_stage}

As illustrated in Figure \ref{fig:model_stages} the neural network stage used during this study was a 3 layer architecture, consisting of an input stage with 30000 inputs, a hidden layer with 40 hidden units and an output layer with 4 units.  The input and hidden layers also possessed a bias unit. The hidden layer used a $tanh$ non-linearity activation function, while the output layer possessed a linear activation function.  The weights of both the hidden and output layers were initialised using random values between -0.05 and 0.05.  This architecture was selected following a hyperparameter sweep  that produced optimal validation set performance using 40 training and 13 validation samples.

Once the network architecture was selected, leave-one-out cross validation was used over the remaining 198 sample test data set to evaluate the overall performance of the model.

\subsection{Forward \& Back Propagation}
\label{sec:forward_prop}  

During training, the forward propagation process differed from the standard method \cite{LeCun1998} used for a neural network slightly in that the hidden unit layer also incorporated the use of a 70\% dropout stage to improve generalisation performance \cite{Baldi2014}.  The contribution of each weight within the network to the overall error was computed by backpropagating the sum of squares error function result from the output layer through the network \cite{LeCun1998}.

\subsection{Weight Updates}
\label{sec:weight_updates}

Following backpropagation, the weights at each of the layers of the neural network were updated to reflect their respective contributions to the error.  For the output layer using a linear activation function, the optimal weights were determined using the Moore-Penrose pseudo-inverse \cite{DeChazal2015}.

The updates to the hidden layer weights were calculated using the standard update method with momentum.  The learning rate used to update the weights; $\alpha$ started at 0.002 and was reduced by 5\% for each improvement in network performance.  The momentum $\mu$ was fixed at 0.95 throughout training.

\subsection{Determining Predicted Mask Size}
\label{sec:determining_mask_size}

The leave-one-out training \& test regime was repeated 4 times with network weights re-initialised with random values prior to each repetition.  Thus 4 nasal landmark predictions were produced for each test image. The predicted landmarks were then transferred back into image space by re-applying the mean and maximum values calculated in section \ref{sec:pre-processing}.  The mean position of each landmark was then computed for each of the test set images.  The nose width was determined using the length of the line between the two mean predicted landmarks and the scale of the image as measured in section \ref{sec:image_dataset}.  This nose width estimate was then used with the known mask sizes in Table \ref{tab:eson_sizes} to determine the \textit{Predicted} mask size.

\section{Results}
\label{sec:results}

Tables \ref{tab:pred_confusion_matrix} to \ref{tab:sensitivity_predict_perf} summarise the performance of the neural network sizing model and the manual measurements taken from the test images; while 
Figures \ref{fig:landmark_predn_performance}a \& b provide examples of accurate and inaccurate landmark predictions. Using the confusion matrix in Table \ref{tab:pred_confusion_matrix}, the semi-automatic sizing system achieved an overall accuracy of  72\% in correctly estimating the mask size and achieved a within one size accuracy of 96\%.  The within one size accuracy figure states that the system predicted the correct size or an adjacent size for 96\% of samples. The manually measured accuracy metrics of 89\% and 100\% respectively are shown in Table \ref{tab:manual_confusion_matrix}.

\begin{table}[!t]
\centering
\renewcommand{\arraystretch}{1.4}
\begin{tabular}{cc  c | c | c | c |}

	& & \multicolumn{4}{c }{\textbf{Predicted Mask Sizes}} \\
	\cline{3-6}
	\multirow{6}{*}{\rotatebox[origin=l]{90}{\textbf{Actual Mask Size}}} & & \multicolumn{1}{|c |}{Small} & \multicolumn{1}{c |}{Medium} & \multicolumn{1}{c |}{Large} & \multicolumn{1}{c |}{Too Large}\\
	\cline{2-6}
	& \multicolumn{1}{|c |}{Small} & 61 & 26 & 5 & 1\\
	\cline{2-6}
	\multicolumn{1}{ c}{}
	& \multicolumn{1}{|c |}{Medium} & 8 & 55 & 9 & 0\\
	\cline{2-6}
	\multicolumn{1}{ c}{}
	& \multicolumn{1}{|c |}{Large} & 1 & 3 & 21 & 1\\
	\cline{2-6}
	\multicolumn{1}{ c}{}
	& \multicolumn{1}{|c |}{Too Large} & 0 & 1 & 0 & 6\\
	\cline{2-6}
\end{tabular}
\caption{Confusion matrix of predicted mask sizes (using the nose width from the mean test set landmarks)}
\label{tab:pred_confusion_matrix}
\end{table}

\begin{table}[!t]
\centering
\renewcommand{\arraystretch}{1.4}
\begin{tabular}{cc  c | c | c | c |}

	& & \multicolumn{4}{c }{\textbf{Manually Measured Mask Sizes}} \\
	\cline{3-6}
	\multirow{6}{*}{\rotatebox[origin=l]{90}{\textbf{Actual Mask Size}}} & & \multicolumn{1}{|c |}{Small} & \multicolumn{1}{c |}{Medium} & \multicolumn{1}{c |}{Large} & \multicolumn{1}{c |}{Too Large}\\
	\cline{2-6}
	& \multicolumn{1}{|c |}{Small} & 81 & 12 & 0 & 0\\
	\cline{2-6}
	\multicolumn{1}{ c}{}
	& \multicolumn{1}{|c |}{Medium} & 2 & 66 & 4 & 0\\
	\cline{2-6}
	\multicolumn{1}{ c}{}
	& \multicolumn{1}{|c |}{Large} & 0 & 1 & 23 & 2\\
	\cline{2-6}
	\multicolumn{1}{ c}{}
	& \multicolumn{1}{|c |}{Too Large} & 0 & 0 & 0 & 7\\
	\cline{2-6}
\end{tabular}

\caption{Confusion matrix of mask sizes manually measured from images}
\label{tab:manual_confusion_matrix}
\end{table}

\begin{table}[b]
\centering
\renewcommand{\arraystretch}{1.4}
\begin{tabular}{l |c|c|c|c|c|c|c|c|}
	\cline{2-9}
	& \multicolumn{4}{|c}{\textbf{Sensitivity (\%)}} & \multicolumn{4}{|c|}{\textbf{Pos. Predict. (\%)}} \\
	\cline{2-9}
	& S & M & L & TL & S & M & L & TL\\
	\hline
	\multicolumn{1}{|c|}{\rotatebox[origin=c]{90}{\textbf{\thinspace Pred \thinspace}}} & 66 & 76 & 81 & 86 & 87 & 65 & 60 & 75\\
	\hline
	\multicolumn{1}{|c|}{\rotatebox[origin=c]{90}{\textbf{\thinspace Man \thinspace}}} & 87 & 92 & 88 & 100 & 98 & 84 & 85 & 78\\
	\hline
\end{tabular}
\caption{Sensitivity \& positive predictivity performance of predicted (Pred) mask sizes and those determined from manual photograph (Man) measurements (S: Small, M: Medium, L: Large, TL: Too Large)}
\label{tab:sensitivity_predict_perf}
\end{table}

\section{Discussion}
\label{sec:discussion}

The semi-automated mask sizing system produced an accuracy of 72\%, while the manual measurements taken from the photograph achieved 89\%.  It was expected that the manual annotator would achieve 100\% accuracy.  A review of the images revealed that parallax error contributed to the error of the system. The nasal region in some images was also obscured thus making landmarking more difficult.  Such an issue could contribute significantly to the measurement / prediction error through an incorrect identification of nasal width, particularly affecting those measurements which are very close to the 2\% size tolerance.

Another potential source of error could result from the location of the 20 cent piece on the forehead.  It is possible that in some images the reference coin is some distance away from the widest part of the nose in terms of depth.  This difference in depth could introduce error as the physical distance represented by a single pixel at nose and coin would differ; thus producing errors in final sizing.  The model could benefit from additional work investigating more reliable means of defining scale within the images.

\begin{figure}[!h]
\begin{minipage}[b]{0.48\linewidth}
  \centering
  \centerline{\includegraphics[height=2.5cm]{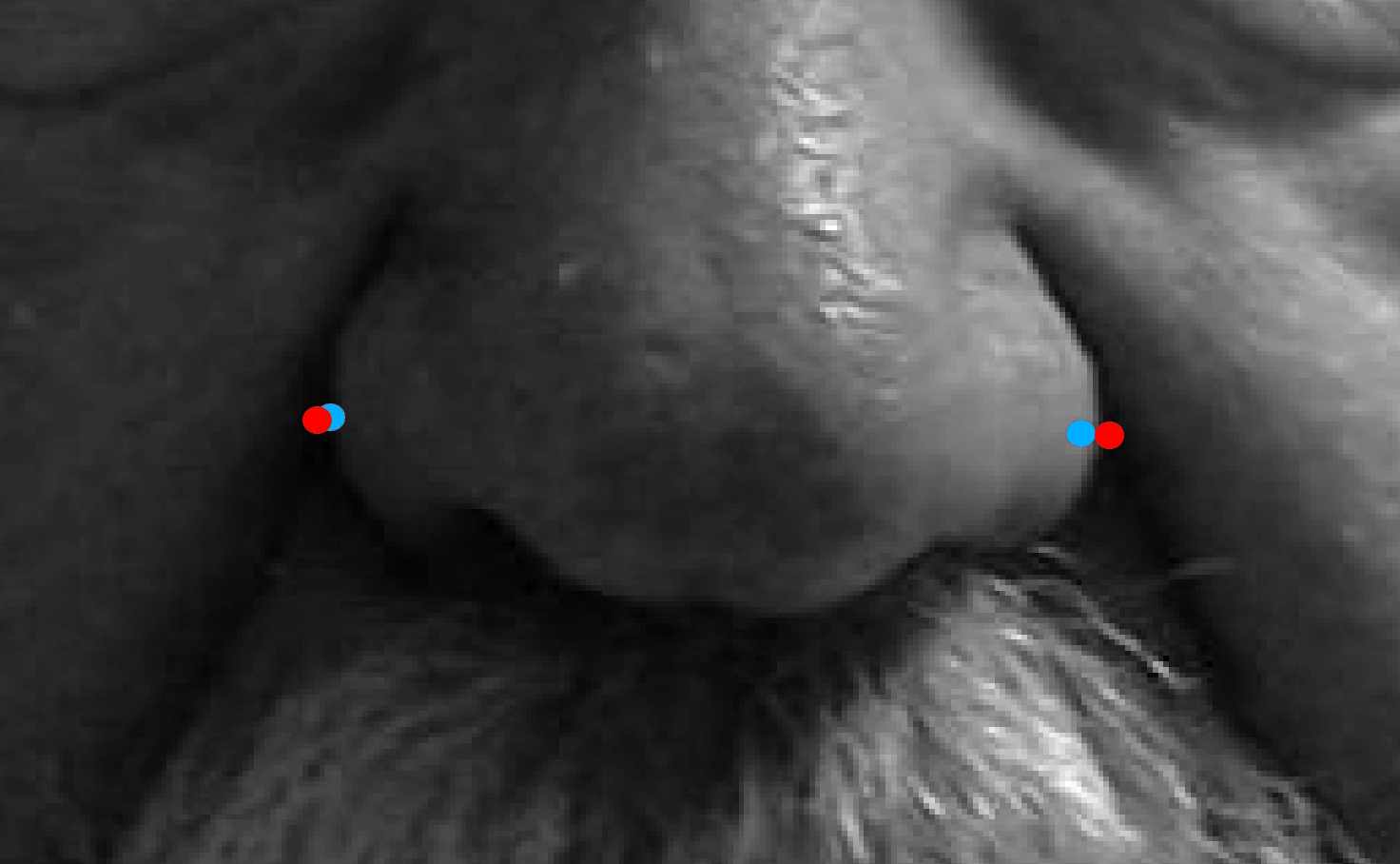}}
  \centerline{(a) Correct prediction}\medskip
  
\end{minipage}
\hfill
\begin{minipage}[b]{0.48\linewidth}
  \centering
  \centerline{\includegraphics[height=2.5cm]{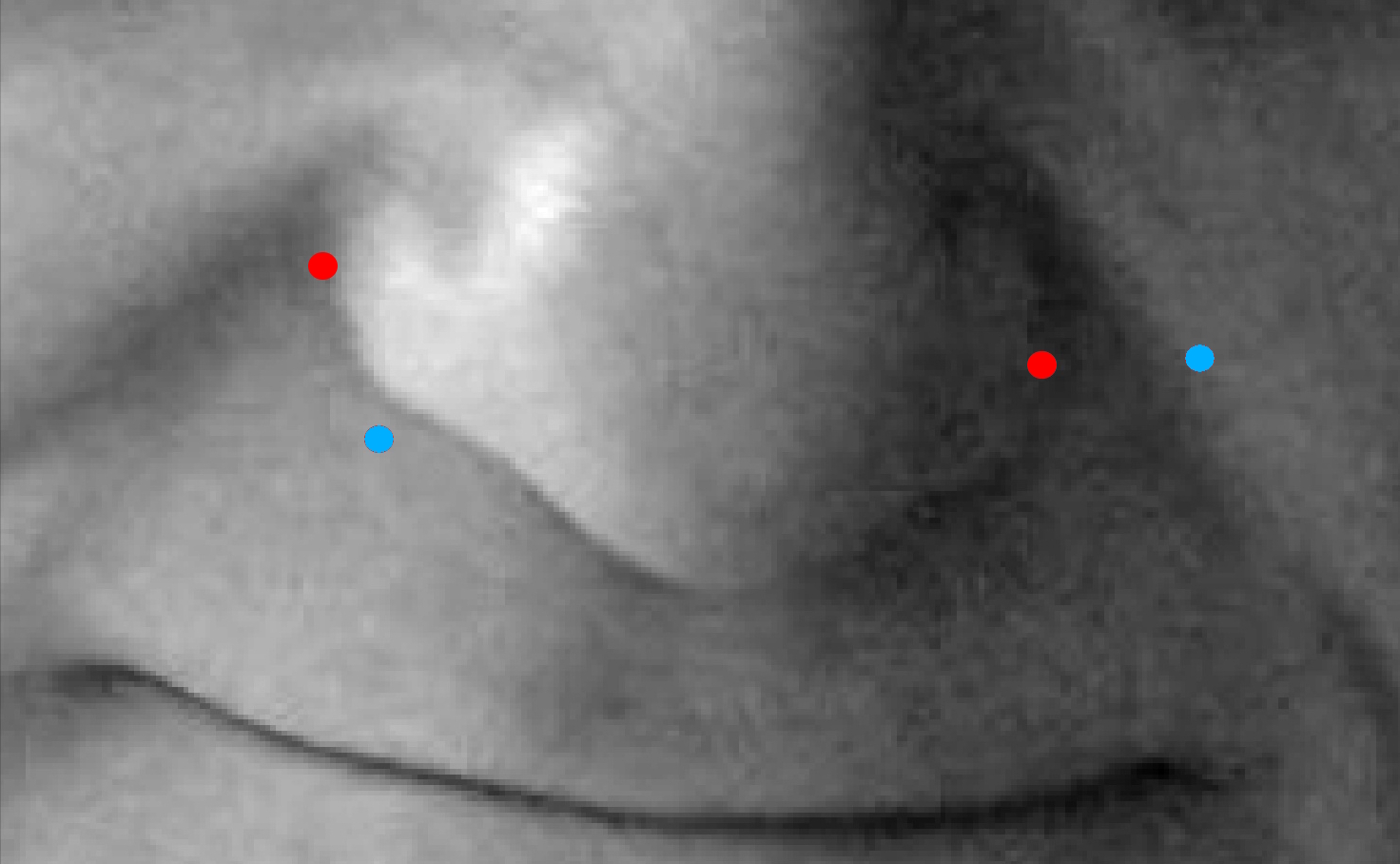}}
  \centerline{(b) Inaccurate prediction}\medskip
\end{minipage}

\caption{Nasal landmark prediction examples (blue dots: predicted landmarks, red dots: labelled landmarks)}
\label{fig:landmark_predn_performance}

\end{figure}

Given the modest sample size of the datasets used in this study the system would almost certainly benefit from additional data.  This is particularly evident examining the images where the system performed least effectively (e.g. Figure \ref{fig:landmark_predn_performance}b); by increasing the number and variety of nose types within the data we anticipate that the overall performance of the system would improve.

\section{Conclusion}
\label{sec:conclusion}
We intend to continue to improve the accuracy of the proposed system: collecting a larger data set, investigating the use of more advanced techniques such as convolutional neural networks to improve facial landmarking accuracy as well as methods to offset errors within the system.  In order for a future version of the system to be effectively used in the field, a means of automatically determining scale within the image must also be developed.

This study presented a semi-automated nasal PAP mask sizing system based on neural networks.  The system achieved 72\% accuracy in correctly estimating the Fisher \& Paykel Eson mask size and a 96\% accuracy in sizing within one size.  This level of performance was comparable to that achieved using manual measurements on the same data set.  Further improvements could be made increasing and refining the data sets used to train and evaluate the model as well as investigating additional methods of determining scale.

\bibliographystyle{IEEEbib}

\end{document}